\pdfoutput=1
\PassOptionsToPackage{table,xcdraw,usenames,dvipsnames}{xcolor}
\documentclass[11pt]{article}
\usepackage[final]{acl}
\usepackage{times}
\usepackage{latexsym}

\usepackage[T1]{fontenc}

\usepackage[utf8]{inputenc}

\usepackage{microtype}
\usepackage{graphicx}
\usepackage{subfigure}
\usepackage{booktabs} 
\usepackage{comment}
\usepackage{hyperref}
\usepackage{algorithm}
\usepackage{algpseudocode}

\usepackage[dvipsnames]{xcolor}
\PassOptionsToPackage{table,xcdraw,usenames,dvipsnames}{xcolor}
\usepackage{microtype}
\usepackage{graphicx}
\usepackage{subfigure}
\usepackage{booktabs} 
\usepackage{hyperref}

\usepackage{amsmath}
\usepackage{amssymb}
\usepackage{mathtools}
\usepackage{amsthm}

\usepackage[capitalize,noabbrev]{cleveref}
\usepackage[ruled,algo2e]{algorithm2e}

\usepackage{multirow}
\usepackage{makecell}
\usepackage{enumerate}
\usepackage{enumitem}
\usepackage{pifont}

\usepackage{xspace}
\usepackage{xspace}     
\usepackage{amsmath}    
\usepackage[utf8]{inputenc} 
\usepackage[T1]{fontenc}    
\usepackage{hyperref}       
\usepackage{url}            
\usepackage{booktabs}       
\usepackage{amsfonts}       
\usepackage{nicefrac}       
\usepackage{microtype}      
\usepackage{xcolor}         
\usepackage{xspace}

\usepackage{microtype}
\usepackage{hyperref}
\usepackage{url}
\usepackage{booktabs}
\usepackage[]{algorithm2e}
\SetKwInput{KwInput}{Input}
\SetKwInput{KwOutput}{Output}

\usepackage{amsmath}

\usepackage{lineno}

\usepackage[textsize=tiny]{todonotes}

\usepackage[normalem]{ulem}
\usepackage{bbm}
\usepackage{bbding}
\usepackage{xspace}

\usepackage{amsmath,amsfonts,bm}

\def\eqref#1{equation~\ref{#1}}

\def\1{\bm{1}}

\DeclareMathAlphabet{\mathsfit}{\encodingdefault}{\sfdefault}{m}{sl}
\SetMathAlphabet{\mathsfit}{bold}{\encodingdefault}{\sfdefault}{bx}{n}

\DeclareMathOperator*{\argmax}{arg\,max}

\SetKwInOut{Input}{Input}\SetKwInOut{Output}{Output}
\theoremstyle{plain}

\theoremstyle{definition}

\theoremstyle{remark}

\ifx\figurename\undefined \def\figurename{Figure}\fi
\renewcommand{\figurename}{Fig.}
\renewcommand{\paragraph}[1]{\textbf{#1} }

\newcommand{\proj}{\textsc{AnyMAC}\xspace}
\newcommand{\mode}[1]{\underline{\textsc{#1}}\xspace}

\newcommand{\RNum}[1]{\uppercase\expandafter{\romannumeral #1\relax}}

\definecolor{ForestGreen}{RGB}{34,139,34}
\definecolor{myyellow}{RGB}{181, 181, 27}

\newcommand{\blue}[1]{$_{\textcolor{BlueGreen}{(- #1)}}$}
\newcommand{\red}[1]{$_{\textcolor{RedOrange}{(+ #1)}}$}

\definecolor{darksalmon}{rgb}{0.91, 0.59, 0.48}
\definecolor{emerald}{rgb}{0.31, 0.78, 0.47}
\definecolor{green(pigment)}{rgb}{0.0, 0.65, 0.31}
\definecolor{amaranth}{rgb}{0.9, 0.17, 0.31}
\definecolor{iris}{rgb}{0.35, 0.31, 0.81}
\definecolor{uu}{rgb}{0.95, 0.51, 0.51}
\definecolor{spirodiscoball}{rgb}{0.06, 0.75, 0.99}

\graphicspath{{figs/}}

\title{AnyMAC: Cascading Flexible \underline{M}ulti-\underline{A}gent \underline{C}ollaboration \\ via Next-Agent Prediction}

\author{
  Song Wang\textsuperscript{\ding{171}},  
  Zhen Tan\textsuperscript{\ding{168}},  
  Zihan Chen\textsuperscript{\ding{171}},  
  Shuang Zhou\textsuperscript{\ding{169}},  
  Tianlong Chen\textsuperscript{\ding{170}},  
  Jundong Li\textsuperscript{\ding{171}} \\
  \textsuperscript{\ding{171}}University of Virginia,
  \textsuperscript{\ding{168}}Arizona State University\\
  \textsuperscript{\ding{169}}University of Minnesota Twin Cities,
  \textsuperscript{\ding{170}}University of North Carolina at Chapel Hill\\
  {\tt \{sw3wv,brf3rx,jundong\}@virginia.edu},\quad {\tt ztan36@asu.edu} \\
  {\tt zhou2219@umn.edu},\quad {\tt tianlong@cs.unc.edu}
}

\begin{document}
\maketitle

\begin{abstract}
Recent progress in large language model (LLM)-based multi-agent collaboration highlights the power of structured communication in enabling collective intelligence. However, existing methods largely rely on static or graph-based inter-agent topologies, lacking the potential adaptability and flexibility in communication.
In this work, we propose a new framework that rethinks multi-agent coordination through a sequential structure rather than a graph structure, offering a significantly larger topology space for multi-agent communication. Our method focuses on two key directions: (1) Next-Agent Prediction, which selects the most suitable agent role at each step, and (2) Next-Context Selection (NCS), which enables each agent to selectively access relevant information from any previous step. Together, these components construct task-adaptive communication pipelines that support both role flexibility and global information flow. Extensive evaluations across multiple benchmarks demonstrate that our approach achieves superior performance while substantially reducing communication overhead. Our code is provided at \href{https://github.com/SongW-SW/AnyMAC}{https://github.com/SongW-SW/AnyMAC}. 

\end{abstract}

\section{Introduction}

The rise of large language models (LLMs) has revolutionized many domains by enabling powerful agents that can perform complex reasoning, planning, and action execution~\citep{self-correction,meta-gpt,zhuge2024gptswarm,tan2024large, li2025visual,Wang2025ReKnoS}. These LLM-based agents, which integrate language generation with decision-making and external tool use, have demonstrated remarkable capabilities in diverse tasks, such as chain-of-thought reasoning~\citep{yao2023react,wang2023selfconsistency} and code synthesis~\citep{reflexion,chen2023codet}. Beyond single-agent settings, recent work has shown that teams of LLM agents can collaboratively solve harder problems than any individual agent~\citep{arXiv2023_MultiAgent-Debate, multi-persona, reflexion, PHPrompting, autogen, zhang2023exploring,qu2025efficient,lei2025learning}, giving rise to an emergent form of collective intelligence. 
This emergent capability hinges critically on the design of inter-agent communication topologies: how agents are structured, how they exchange messages, and how they integrate information from others.

\begin{figure}[t]
\begin{center}
\begin{tabular}{c}
\includegraphics[width = 0.99\linewidth]{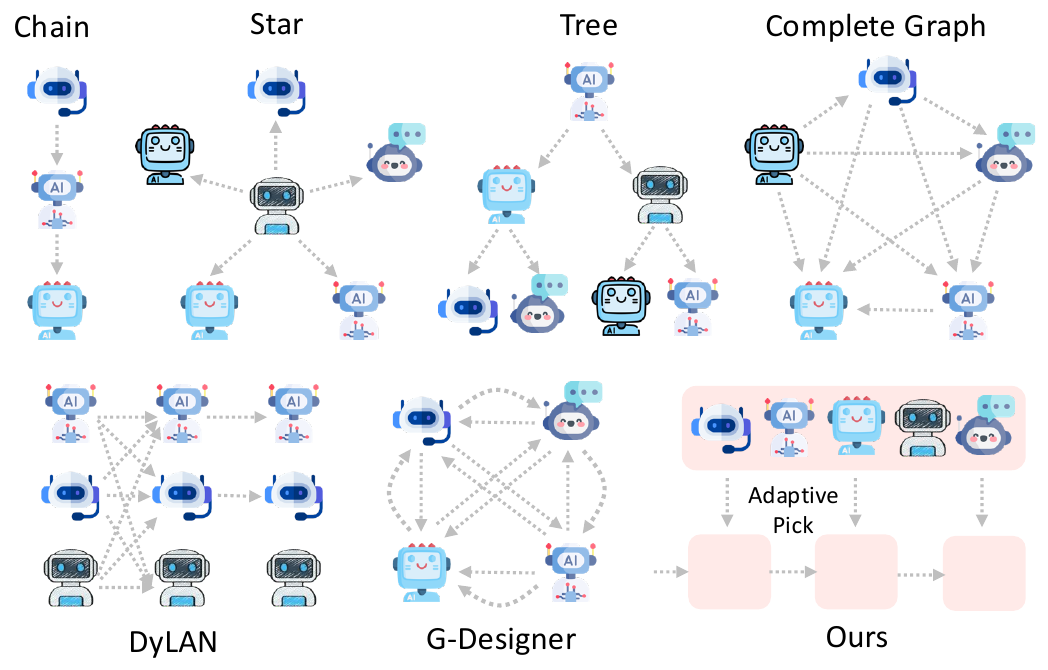}
\end{tabular}
\end{center}
\caption{Comparison of LLM-based multi-agent communication topology design.}
\label{Fig1}
\vspace{-0.2cm}
\end{figure}

To support such collaboration, researchers have investigated a wide range of multi-agent communication structures (Fig.~\ref{Fig1}), including chains~\citep{cot,zhang2022automatic}, trees~\citep{tot}, stars~\citep{autogen}, fully connected or random graphs~\citep{qian2024scaling}, and learned or optimizable topologies~\citep{zhuge2024gptswarm,zhang2024cut}. These designs, often tailored to task complexity or communication budgets, aim to balance performance and efficiency in various deployment scenarios. Notably, recent approaches have introduced learning-based topology construction~\citep{chatllm-network, arXiv2023_Dynamic-LLM-Agent,zhang2024g}, enabling dynamic selection of agent communication graphs conditioned on input tasks and queries. Such adaptive frameworks mark a shift from fixed pipelines to more flexible, input-aware systems that can better exploit the potential of LLM collectives.

Despite these advancements, current graph-based structures still face fundamental limitations. First, they enforce \emph{static communication schemas} within each round: once the topology is learned, all agents operate under the same fixed communication pattern, preventing the reuse of agents or dynamic adaptation during the reasoning process. Additionally, to maintain acyclic message flow, many designs restrict the graph to be a Directed Acyclic Graph (DAG), which further constrains the solution space (of communication topology) and prohibits recursive or repeated consultation of specific agents. 
For example, in a task where one expert agent (e.g., a Python coder) is particularly useful at multiple stages of reasoning, a DAG-based structure cannot re-query this agent after it is used earlier in the round. This leads to inefficient or suboptimal reasoning, especially in complex tasks where revisiting agents is crucial.
Second, most existing works limit information flow strictly to direct graph edges between agents, meaning that each agent can only access messages from its neighbors. 
%
%
%
For example, in tree-based structures, downstream agents often lack access to parallel branches’ outputs, missing potentially useful contextual signals. This makes it hard for the agents to obtain global context for well-informed reasoning.
%

To address these challenges, we propose a new multi-agent collaboration framework, namely \proj, that formulates multi-agent collaboration through a \emph{sequential communication protocol} rather than a graph-based one. 
In this way, the construction of the communication topology is formulated as predicting the next agent iteratively.  
Our framework contains two novel and critical designs:
(1) \textbf{Next-Agent Prediction}, where the system dynamically determines the next agent to activate in a stepwise manner. 
This sequential design bypasses the constraints of graph structures, allowing for greater flexibility in agent reuse and order variation across different queries. 
(2) \textbf{Next-Context Selection}, which allows each step to flexibly retrieve outputs from any previously activated agents. 
This globally accessible mechanism enables richer and more adaptive communication flows, where information is not constrained to propagate through fixed graph edges or sequential orders, but instead can be retrieved through dynamic selection based on task requirements. 
We conduct extensive experiments across multiple benchmarks, and the results validate the effectiveness of our approach, outperforming state-of-the-art communication topologies in both accuracy and efficiency in terms of token consumption. 
Our contributions can be summarized as follows:

\begin{itemize}[leftmargin=*]
    \item \textbf{Formulation.} We propose a new formulation of multi-agent communication, where the system predicts the next agent role and selects context from any previous agents. This formulation is proven to subsume the solution space of prior graph-based methods.

    \item \textbf{Framework.} We propose a transformer-based framework to realize our formulation, leveraging the transformer's global attention and sequential modeling capabilities. 
    
    \item \textbf{Experiments.} We conduct extensive experiments across diverse benchmarks. Our method  outperforms state-of-the-art multi-agent baselines in both accuracy and efficiency, demonstrating adaptivity, robustness, and favorable cost-performance trade-offs.

\end{itemize}






\section{Related Work}
\label{sec:related}

\paragraph{Single Agent Reasoning.}
Recent research has demonstrated that
multi-step reasoning allows large language models (LLMs) to solve complex problems and self-correct along the way. 
Broadly, single agent multi-step reasoning can be achieved via \textit{training-based} and \textit{prompting-based} methods.

In training-based approaches, reinforcement learning (RL) is used to optimize the model's ability to generate long-form Chain-of-Thought (CoT) reasoning~\citep{deepseekr1}. 
While effective, RL methods typically require substantial data and computational resources. 
To reduce cost, distillation-based methods~\citep{muennighoff2025s1, ye2025limo} collect high-quality reasoning traces and apply supervised fine-tuning to teach models multi-step reasoning behaviors.

Beyond training-based methods, prompting-based techniques enable step-by-step reasoning by prompting procedure. 
Early approaches include multi-step reasoning exemplars directly in the prompt~\citep{cot,zhang2022automatic}, or resort to external knowledge with in-context learning~\citep{chen2025maple,wang2024mixture,chen2024fastgas, liu2024knowledge}. Some recent efforts also consider the reasoning's trustworthiness~\cite{tan2024glue,tan2025tuning} and explainability~\cite{manuvinakurike2025thoughts,tan2024interpreting, hu2024understanding}.
Previous work also explicitly enforces multi-step reasoning in prompting procedure~\cite{tan2025prospect}, such as ToT~\citep{tot} and budget-forcing~\citep{muennighoff2025s1}.
Beyond single-agent reasoning, \textit{multi-agent} approaches leverage collaboration among multiple LLMs to further improve accuracy, detailed below.

\noindent\textbf{Multi-Agent Collaboration.}
Existing multi-agent systems typically predefine role types and fix the number of agents per role based on the task, then design a communication topology for collaboration. Prior work can be categorized by how this topology is generated.
(1) Early approaches adopt \textit{static} structures, such as chain~\citep{software-dev, meta-gpt, holt2024l2mac}, star~\citep{autogen, yan2024depending, zhou2023large}, and tree~\citep{ishibashi2024selforganize-mother}, which remain unchanged across tasks.
(2) To improve adaptability, recent methods have explored learning static communication graphs from data~\cite{zhang2025symbiotic}. 
GPTSwarm~\citep{zhuge2024gptswarm} parameterizes agent interactions using predefined Directed Acyclic Graph (DAG) topologies and optimizes them using reinforcement learning. 
However, the resulting structures remain fixed across the dataset and are input-independent, lacking the flexibility to adapt communication to individual task instances.
(3) Recent efforts explore query-adaptive topology generation, such as DyLAN~\citep{arXiv2023_Dynamic-LLM-Agent} and G-Designer~\citep{zhang2024g}, where agent interactions are dynamically constructed based on the input. While more flexible, they still rely on a manually defined number of agents and are constrained by canonical graph structures (i.e., the anchor structure).
%
%
In contrast, our formulation allows the network to adaptively determine both the number of agents and the communication structure, without being restricted by a predefined topology. 
This flexibility enables exploration of a significantly larger topology space, leading to more effective and adaptive collaboration.

\section{Problem Formulation}

\label{sec:formulation}


\begin{figure*}
\begin{center}
\includegraphics[width = 0.95\linewidth]{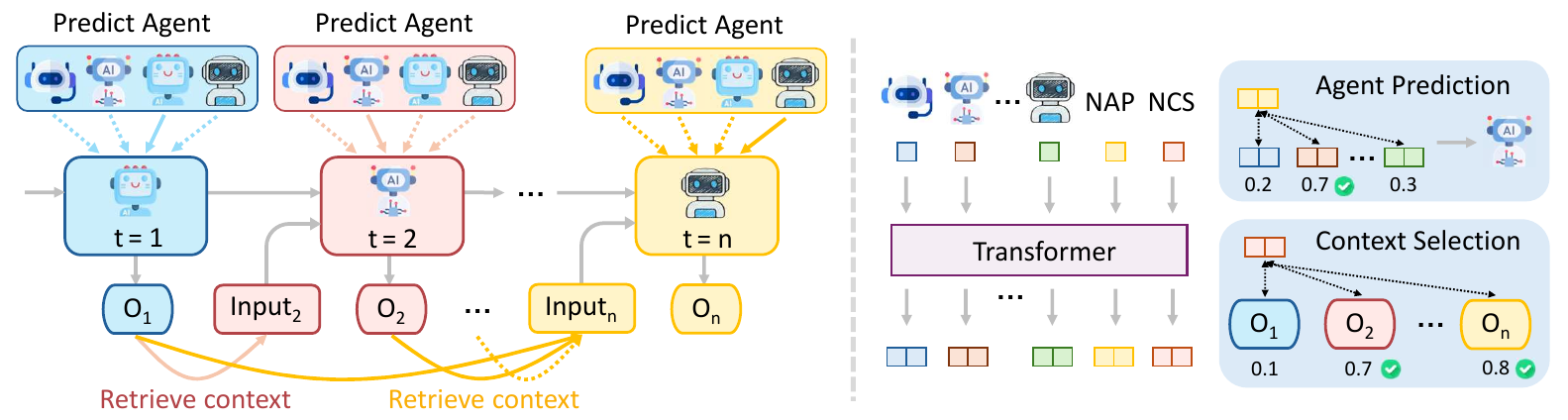}
\end{center}
\caption{The overview of our proposed framework \proj. Left-hand side: At each time step, we perform two stages of operations: (1) Next-Agent Prediction (NAP), which aims to select the most suitable agent role from a set of candidate roles. (2) Next-Context Selection (NCS), which aims to retrieve useful context from the outputs of previously activated agents. The retrieved context will act as the input to the selected agent. Right-hand side: Given the embeddings of a series of activated agents, we perform contextual encoding using a transformer-based model to encode them with additional NAP and NCS tokens. The output embeddings of NAP and NCS tokens will be used to select the next agent and retrieve context from the next agent, respectively.}
\label{Fig2_pipeline}
\end{figure*}

In this section, we define the key concepts for our sequential agent collaboration framework.
Unlike prior works that formulate the multi-agent topology as a fixed directed acyclic graph (DAG), we represent the communication pipeline as a \textbf{sequence} $\mathcal{S} = [a_1, a_2, \dots, a_T]$, where each element $a_t$ is an LLM-based agent selected at the $t$-th step. This design allows agents to be reused multiple times and enables dynamic adjustment of the interaction order based on the task.

Each agent $a_t$ is defined by:
\begin{equation}
a_t = \{\texttt{Base}_t, \texttt{Role}_t, \texttt{State}_t, \texttt{Tool}_t\},
\end{equation}
where $\texttt{Base}_t$ is the underlying language model instance, $\texttt{Role}_t$ indicates the agent's role, $\texttt{State}_t$ captures its memory and interaction history, and $\texttt{Tool}_t$ is an optional set of plugins (e.g., calculator, search engine, or file retriever).

Given an initial query $\mathcal{Q}$, the communication sequence unfolds over $T$ steps. At each step $t$, the system predicts the next agent $a_t$ and composes a prompt $\mathcal{P}^{(t)}$ containing both the original query and selected messages from previous steps:
\begin{equation}
\mathcal{P}^{(t)}_R = \texttt{Select}\left( \{\mathcal{O}^{(1)}, \dots, \mathcal{O}^{(t-1)}\} \right),
\end{equation}
where $\mathcal{O}^{(t-1)}$ denotes the response generated by agent $a_{t-1}$, and $\texttt{Select}(\cdot)$ is a learnable module that chooses relevant past outputs to include in the current prompt. This enables \textbf{flexible and global context access}, unlike graph-based strategies that are restricted to local neighborhoods.

Each agent executes based on its own system and user prompt:
\begin{equation}
\mathcal{O}^{(t)} = a_t\left( \mathcal{P}_\text{sys}^{(t)}, \mathcal{P}_\text{usr}^{(t)}, \mathcal{P}^{(t)}_R \right),
\end{equation}
where $\mathcal{P}_\text{sys}^{(t)}$ includes $\texttt{Base}_t$, $\texttt{Role}_t$, $\texttt{State}_t$, and $\texttt{Tool}_t$, and $\mathcal{P}_\text{usr}^{(t)}$ is the user prompt, which may include the query and task instructions from the user. After $T$ steps, a final referee agent will aggregate the output and provide the final answer.

\section{Methodology}\label{sec:method}
As introduced in \Cref{sec:formulation}, we formulate the problem of LLM-based multi-agent collaboration as a sequential decision process. Our framework, \proj, dynamically constructs a communication sequence $\mathcal{S} = [a_1, a_2, \dots, a_T]$ by predicting, at each step, the next agent $a_t$ and the relevant context to be passed as input. This formulation overcomes the rigidity of fixed graph topologies by allowing agent reuse and flexible context routing across steps.

\Cref{Fig2_pipeline} illustrates the workflow of \proj. Given a task query $\mathcal{Q}$, a set of candidate agent roles $\mathcal{R}$, and an optional tool set, our model iteratively builds the communication sequence. At each step $t$, it performs three stages: \textbf{Encoding}, \textbf{Prediction}, and \textbf{Execution}. 
In the \textit{Encoding} stage, the $\mathcal{Q}$, $\mathcal{O}$, and historical conversation $\mathcal{H}_{t-1}$  are tokenized. These tokens are then fed into a Transformer to obtain contextual embeddings.
In the \textit{Prediction} stage, the contextual embeddings are used for the Next Agent Prediction (NAP) and Next Context Selection (NCS).
In the \textit{Execution} stage, we invoke the selected agent (LLM) with the chosen role and context to generate the response. 
The response $\mathcal{O}^{(t)}$ is appended to $\mathcal{H}_t$ and used in the next round until a final aggregation step produces the answer $a^{(T)}$.

\subsection{Contextual Encoding}
\label{sec:prelim}
\paragraph{Semantic Tokenization.} 
At time step $t$, given a task query $\mathcal{Q}$, a set of candidate role descriptions $\mathcal{R}_i,i=1,2,\dotsc, N$, and conversation history $\mathcal{H}_{t-1}$ of previous steps, we begin by encoding these components into embeddings. 
Specifically, the task query $\mathcal{Q}$ contains textual instructions describing the question. 
Each agent role $\mathcal{R}_i$ includes a role prompt that instructs the agent to act in a specific role and provides an optional list of tools the agent can access. 
Each historical conversation in $\mathcal{H}$ corresponds to a previous agent, consisting of the role description of this agent and its associated response. 
%
Let $\texttt{Embed}(\cdot)$ denote an encoder function that outputs an embedding for any input text. Formally, the tokenization process is:
\begin{align}
\mathbf{q} &= \operatorname{\texttt{Embed}}(\mathcal{Q}) , \\
\mathbf{r}_i &= \operatorname{\texttt{Embed}}(\mathcal{R}_i), \ \ i=1,2,\dotsc, N\\
\mathbf{h}^{(j)} &= \mathbf{r}^{(j)} \Vert \operatorname{\texttt{Embed}}(\mathcal{O}^{(j)}),\ \ j=1,2,\dotsc, t-1
\end{align}
Here, $\mathbf{q}$ is the embedding of the task query, $\mathbf{r}_i$ is the embedding of the $i$-th agent role description, and $\mathbf{h}^{(j)}$ is the embedding of the $j$-th historical conversation, obtained by concatenating the role embeddings $\mathbf{r}^{(j)}$ and response embedding $\texttt{Embed}(\mathcal{O}^{(j)})$ of agent $a^{(j)}$. Here we use $a^{(j)}$ to denote the specific agent used at step $j$.
All embeddings are passed through a linear projection layer to match the transformer’s input dimension. 
We use three separate projection layers: one for the task query, one for the role descriptions, and one shared across all historical conversations:
\begin{align}
\tilde{\mathbf{q}} &= f_q(\mathbf{q}), \quad 
\tilde{\mathbf{r}}_i = f_r(\mathbf{r}_i), \quad 
\tilde{\mathbf{h}}_j = f_h(\mathbf{h}_j),
\end{align}
where \( f_q, f_r, f_h \) are learnable linear projections.
Moreover, to enable task-adaptive NAP and NCS, we generate the NAP and NCS tokens $\mathbf{t}_{\text{NAP}}$ and $\mathbf{t}_{\text{NCS}}$ using the task query embedding $\mathbf{q}$ by passing it through two separate linear layers:
\begin{align}
\mathbf{t}_{\text{NAP}} = f_{\text{NAP}}(\mathbf{q}), \quad
\mathbf{t}_{\text{NCS}} = f_{\text{NCS}}(\mathbf{q}),
\end{align}
where \( f_{\text{NAP}} \) and \( f_{\text{NCS}} \) are learnable linear projections that map the query embedding to the input dimension expected by the Transformer.

\noindent\textbf{Contextual Encoding.}
After obtaining the token embeddings, we concatenate them into a sequence 
and feed it into an $L$‑layer Transformer encoder to obtain contextualized embeddings $\mathbf{T}$:
\begin{equation}
\mathbf{T}^\star = \operatorname{Trans}_L\left([\tilde{\mathbf{q}}, \tilde{\mathbf{r}}_{1:N}, \tilde{\mathbf{h}}_{1:t-1}, \mathbf{t}_{\text{NAP}}, \mathbf{t}_{\text{NCS}}]\right).
\end{equation}
This encoding fuses information from task intent, role priors, and dialogue history, producing contextualized embeddings \( \mathbf{T}^\star \) of the same length, which are used in the subsequent NAP and NCS. In the following, we use notations with a superscript of $^\star$ to denote that they are from $\mathbf{T}^\star$.

\subsection{Next-Agent Prediction (NAP)}\label{sec:nap}
\label{nap}
Let $N$ denote the total number of candidate roles.
To determine the next agent role, we first compute pairwise compatibility scores between the contextualized NAP token $\mathbf{t}_{\text{NAP}}^\star$ and each contextualized role token $\mathbf{r}_i^\star$ (both are from $\mathbf{T}^\star$) using an inner product:
\begin{equation}\label{eq:raw-nap}
s_i = \langle \mathbf{t}_{\text{NAP}}^\star, \mathbf{r}_i^\star \rangle, \quad i = 1, \dots, N.
\end{equation}
We then select the role of the next agent (i.e., $a_t$) by considering the role $\mathcal{R}_k$ with the highest compatibility score $s_i$: 
\begin{equation}
   a_t=\mathcal{R}_k,\quad\text{where}\quad  k= \argmax_{i=1,2,\dotsc, N} s_i.
\end{equation}
During training, to encourage exploration, we apply the Gumbel‑Softmax~\cite{jang2016categorical} over $\{s_i\}$.

\subsection{Next-Context Selection (NCS)}\label{sec:NCS} 
\label{ncs}
To decide which parts of the historical conversation should be fed into the next agent as context, we compute cosine similarity scores between the contextualized NCS token $\mathbf{t}_{\text{NCS}}^\star$ and each contextualized historical conversation token $\mathbf{h}_j^\star$ (both from $\mathbf{T}^\star$):
\begin{equation}
c_j =  \frac{\langle \mathbf{t}_{\text{NCS}}^\star, \mathbf{h}_j^\star \rangle}
         {\|\mathbf{t}_{\text{NCS}}^\star\| \cdot \|\mathbf{h}_j^\star\|}, \quad c_j \in [-1, 1],
\end{equation}
These scores are then passed through a sigmoid function, resulting in values that lie in \((0, 1)\):
\begin{equation}
\label{eqn:sigmoid}
g_j = \sigma(c_j) = \frac{1}{1 + e^{-c_j}}, \quad j=1,2,\dotsc, t-1,
\end{equation}
to obtain gate values $g_j$, which determine the inclusion probability for each context.
At inference time, we select the context from previous agents by applying a threshold: 
\begin{equation}
\mathcal{P}^{(t)}_O =\{(\mathcal{R}^{(j)}, \mathcal{O}^{(j)})|g_j\geq \eta,j=1,2,\dotsc, t-1\},
\end{equation}
where $\eta\in\mathbb{R}$ is a threshold used for selecting.
During training, we sample Bernoulli masks with probabilities given by \( g_j \) to enable exploration.


\noindent\textbf{Sparsity Penalty.} Ideally, we would like the agent to access as much historical context as possible. 
However, excessively long context can lead to two issues: the agent may become overwhelmed by irrelevant information, and the API cost increases with longer context. 
To address this, we propose a sparsity penalty during training:
%
\begin{equation}
\mathcal{L}_{\text{sparse}} = \lambda \sum_{j=1}^{t-1} |g_j|.
\end{equation}
Here, $g_j$ is the gating score for the $j$-th historical conversation (i.e., $\mathbf{h}_j^\star$). By penalizing the magnitude of $g_j$, the model learns to selectively include only the most relevant context, avoiding unnecessary prompt tokens.

\subsection{Execution}\label{sec:execution}
Once the next agent role and its associated context are determined, we perform \textit{Execution}. 
We prompt the selected agent $a_t$ with the system prompt $\mathcal{P}^{(t)}_\text{sys}$, the user prompt $\mathcal{P}^{(t)}_\text{usr}$, and selected context $\mathcal{P}_R^{(t)}$, to generate the  response $\mathcal{O}^{(t)}$:
\begin{equation}
\mathcal{O}^{(t)} = a_t(\mathcal{P}^{(t)}_\text{sys}, \mathcal{P}^{(t)}_\text{usr}, \mathcal{P}_O^{(t)}).
\end{equation}
The output is appended to the historical conversation buffer, enabling iterative reasoning and coordination in subsequent rounds. Formally, we update the conversation history as:
\begin{equation}
\mathcal{H}_{t} = \mathcal{H}_{t-1} \cup \{(a_t, \mathcal{O}^{(t)})\},
\end{equation}
where $a_t$ denotes the selected agent based on NAP scores at time step $t$, and $\mathcal{O}^{(t)}$ is the generated response after executing the role $a_t$ using the selected context $\mathcal{P}_O^{(t)}$.
This process continues iteratively until a final decision agent is selected or the maximum number of iterations is reached.


\subsection{Optimizing \proj with RL}\label{sec:rl}
To improve routing quality with awareness of task difficulty and efficiency, we formulate NAP training as a reinforcement learning (RL) problem.

\noindent\textbf{Efficiency-Aware Reward.}
Given a final aggregated answer $a$ from a communication sequence with length $l$, we define the reward $r$ as:
\begin{equation}
r = \gamma^l \cdot \mathbb{I}[\texttt{is\_correct}(a)],
\end{equation}
where $\mathbb{I}[\cdot]$ is the indicator function that returns 1 if the answer $a$ is correct, and 0 otherwise. The exponential decay term $\gamma^l$, where $\gamma \in (0, 1]$, penalizes longer routing paths. 
A smaller \( \gamma \) places greater emphasis on efficiency by punishing large $l$ more.

\noindent\textbf{Difficulty-Aware Advantage Estimation.}
Since questions vary in difficulty, directly comparing rewards across different queries is misleading. 
This is because a high reward may result from an easy question rather than a good routing policy. 
To address this, we apply a standard reward normalization technique~\citep{gu2016continuous, schulman2017proximal}, which takes the average reward of a question into account and compute advantage $A_t$ of each trajectory, as described in \Cref{sec:rw_norm}.
The policy gradient loss is then computed as follows:
\begin{equation}
\label{eqn:pg}
\nabla_\Theta\mathcal{L}_{\text{PG}}
 = -\sum_{t=1}^{T} A_t \nabla_\Theta\log P_\Theta(\tau_t),
\end{equation}
where $P_\Theta(\tau_t)$ denotes the probability of the $t$-th sampled trajectory under the current model parameters $\Theta$. This includes both the NAP selection probability (\Cref{nap}) obtained via Gumbel-Softmax, and the NCS selection probability (\Cref{ncs}) derived from the sigmoid gating function.
The full objective combines the policy gradient and the sparsity regularization term (\Cref{ncs}):
\begin{equation}
\mathcal{L} = \mathcal{L}_{\text{PG}} + \lambda\,\mathcal{L}_{\text{sparse}},
\end{equation}
where $\lambda$ is a hyper-parameter to control the importance of the sparsity regularization term.

\section{Experiments}
\label{experiments}
\vspace{-0.5em}
\subsection{Experimental Setup}
\vspace{-0.4em}

\paragraph{Benchmarks \& Tasks.}
We evaluate \proj across a diverse suite of tasks spanning general reasoning, mathematical problem solving, and code generation. Specifically, we use \texttt{MMLU}~\citep{mmlu} for general reasoning; \texttt{GSM8K}~\citep{arXiv2021_Verifier-Math}, \texttt{MultiArith}~\citep{roy2016solving}, \texttt{SVAMP}~\citep{patel2021nlp}, and \texttt{AQuA}~\citep{ling2017program} for math reasoning; and \texttt{HumanEval}~\citep{human-eval} for code generation.
We follow the standard train/test splits provided by each dataset.

\noindent\paragraph{Variants.}
We provide two variants of our method: {\proj} and {\proj-Eff}.
The first one disables all efficiency-related constraints by setting the sparsity loss weight $\lambda = 0$, the reward decay factor $\gamma = 1$,  allowing the model to focus solely on accuracy.
In contrast, {\proj-Eff} sets $\lambda = 1\mathrm{e}{-3}$, $\gamma = 0.9$, and constrains the context selection to be within the newest 2 responses, encouraging shorter routing trajectories (i.e., fewer agents) and compact context selection. 

\noindent\paragraph{Baselines.}
To ensure a comprehensive comparison of various methods, we compare \proj against both single-agent prompting and multi-agent collaboration frameworks. 
Single-agent baselines include \mode{COT} (Chain-of-Thought)~\citep{cot}, \mode{ComplexCoT}~\citep{fu2022complexity}, \mode{Self-Consistency}~\citep{wang2023selfconsistency}, and \mode{PHP}~\citep{PHPrompting}. 
Multi-agent baselines include topology-based methods, including \mode{Chain}, \mode{Star}, \mode{Tree}~\citep{qian2024scaling}, \mode{Complete Graph}, and \mode{Random Graph}. For learning-based frameworks, we consider \mode{AutoGen}~\citep{autogen}, \mode{LLM-Debate}~\citep{arXiv2023_MultiAgent-Debate}, \mode{DyLAN}~\citep{arXiv2023_Dynamic-LLM-Agent},  \mode{GPTSwarm}~\citep{zhuge2024gptswarm}, and \mode{G-Designer}~\citep{zhang2024g}. 

\begin{table*}[!t]
  \centering
    \caption{
    Accuracy comparison across single-agent and multi-agent baselines. 
    The best results are shown in \textbf{bold}, and the runner-ups are \underline{underlined}. 
    We highlight the accuracy gain (\textcolor{RedOrange}{+} / \textcolor{BlueGreen}{--}) relative to the vanilla baseline.
    }

  \label{tab:main_acc}
  \renewcommand\tabcolsep{5.5pt}
  \renewcommand\arraystretch{1.15}

  \small
  \begin{tabular}{l|ccccccc}
    \Xhline{1.1pt}
    \rowcolor{gray!15}
    \textbf{Method} & \textbf{MMLU} & \textbf{GSM8K} & \textbf{MultiArith} & \textbf{SVAMP} & \textbf{AQuA} & \textbf{HumanEval} & \textbf{Avg.} \\
    \Xhline{1.1pt}
    Vanilla & 82.14 & 85.40 & 93.15 & 87.18 & 70.34 & 71.68 & 81.65\\
    CoT & 82.65\red{0.51} & 87.17\red{1.77} & 94.79\red{1.64} & 88.32\red{1.14} & 73.91\red{3.57} & 75.52\red{3.84} & 83.73\\
    PHP & 83.45\red{1.31} & \underline{95.50}\red{10.10} & {98.10}\red{2.84} & 90.02\red{3.44} & {79.00}\red{8.66} & 82.96\red{11.36} & 88.17\\
    \Xhline{1.1pt}
    \rowcolor{gray!10}Chain & 82.35\red{0.21} & 85.57\red{0.17} & 94.38\red{1.23} & 83.41\blue{3.77} & 70.94\red{0.60} & 80.88\red{9.20} & 82.92\\
    \rowcolor{gray!10}Star & 80.79\blue{1.35} & 85.55\red{0.15} & 93.79\blue{0.64} & 88.09\red{0.91} & 68.57\blue{1.77} & 75.65\red{3.97} & 82.07\\
    \rowcolor{gray!10}Tree & 81.89\blue{0.25} & 84.56\blue{0.84} & 94.60\red{1.45} & 89.25\red{2.07} & 72.84\red{2.50} & 77.38\red{5.70} & 83.42 \\
    \rowcolor{gray!10}Complete Graph & 83.15\red{1.01} & 86.49\red{1.09} & 97.20\red{4.05} & 89.48\red{2.30} & {79.21}\red{8.87} & 83.75\red{12.07} & 86.55\\
    \rowcolor{gray!10}Random Graph & 83.76\red{1.62} & 86.14\red{0.74} & 95.46\red{2.31} & 85.41\blue{1.77} & 74.07\red{3.73} & 82.66\red{10.98} & 84.58 \\\Xhline{1.1pt}
    \rowcolor{gray!20}AutoGen & 82.13\blue{0.01} & 90.06\red{7.92} & 93.80\red{0.65} & 88.44\blue{1.26} & 73.65\red{3.31} & 85.41\red{13.73} & 85.58 \\
    \rowcolor{gray!20}LLM‑Debate & 83.69\red{1.55} & 90.23\red{4.83} & 96.27\red{3.12} & 90.56\red{3.38} & 77.52\red{7.18} & 83.79\red{12.11} & 87.01 \\
    \rowcolor{gray!20}DyLAN & 80.16\blue{1.98} & 88.16\red{2.76} & 94.27\red{1.12} & 87.40\red{0.22} & 74.16\red{3.82} & {89.70}\red{18.02} & 85.64 \\
    \rowcolor{gray!20}GPTSwarm & 83.98\red{1.84} & 89.74\red{4.34} & 97.84\red{4.69} & 86.42\blue{0.76} & 78.16\red{7.82} & 88.49\red{16.81} & 87.32 \\ 
    \rowcolor{gray!20}G‑Designer & \textbf{84.50}\red{2.36} & 95.07\red{9.67} & \underline{98.30}\red{5.15} & \underline{91.85}\red{4.67} & \underline{79.47}\red{9.13} & \underline{89.90}\red{18.22} & \underline{89.84}\\
    \Xhline{1.1pt}
    \rowcolor{cyan!10} \proj & \underline{84.30}\red{2.16} & \textbf{95.66}\red{10.26} & \textbf{99.44}\red{6.29} & \textbf{92.67}\red{5.49} & \textbf{81.50}\red{11.16} & \textbf{90.12}\red{18.44} & \textbf{90.62}\\
    \Xhline{1.1pt}
  \end{tabular}
  \vspace{-.1in}
\end{table*}

\noindent\paragraph{Implementation.}
We implement all agents using the OpenAI model, \texttt{gpt-4-1106-preview}. 
For all single-agent and multi-agent baselines, we follow the official configurations used in {G-Designer}~\citep{zhang2024g}\footnote{Since the official implementations of baselines are not fully open-source, we use results provided in  G-Designer.}, which adopt a temperature of $0$ for single-agent and $1$ for multi-agent methods. 
For sentence embeddings, we use \texttt{all-MiniLM-L6-v2}~\citep{wang2020minilm} as $\texttt{Embed}(\cdot)$ to encode queries, role descriptions, and historical responses into 384-dimensional vectors.

We also make specific design choices in \proj. 
We use a default temperature of $1$, except on multiple-choice benchmarks (\texttt{MMLU} and \texttt{AQuA}), where we set the temperature to $0$ as we find it helps reduce hallucinations. 
The maximum number of agents is fixed to $5$ across all tasks. 
Once this limit is reached, the next agent role is forced to be the decision model, which produces the final answer and terminates the reasoning process.

We use \texttt{GPT-2 Small}~\citep{radford2019language} as the routing model and initialize it with pretrained weights. 
The computational overhead of the routing model is negligible compared to LLMs: it is lightweight (117M parameters) and only predicts two tokens per step (NAP and NCS). 
In contrast, LLMs are over 1000$\times$ larger (e.g., \texttt{GPT-3} has 175B parameters~\citep{brown2020language}) and generate hundreds of tokens per response.
To train the model, we sample 80 questions from each dataset and collect 1000 routing trajectories along with their corresponding rewards.

\noindent\paragraph{Training questions selection.} 
If a dataset provides a train/test split, we sample $80$ task instances from the training set; otherwise, we use the first $80$ tasks in test set. 
The model is optimized separately for each dataset.

\noindent\paragraph{Adaptive Sampling.} 
We collect $1000$ trajectory samples (attempted answers) per dataset.
To make effective use of the training sample budget, we design an adaptive sampling strategy that allocates more samples to difficult questions. 
Specifically, when iterating through all questions in a dataset, we do not set a fixed number of trials per question. 
Instead, we define a required number of correct answers for each question before moving on. 
This strategy naturally allocates more sampling budget to harder examples, as easy questions tend to reach the threshold with fewer samples, while difficult ones require more. 
After each epoch, if there is remaining trajectory budget, we proceed to the next epoch and repeat the process until the total sampling budget is exhausted.
For {\proj}, we set the threshold to $1$ correct answer; for {\proj-Eff}, we increase it to $4$ to encourage more stable training.

\subsection{Comparative Results}

\noindent\textbf{\proj outperforms other baselines.} Table~\ref{tab:main_acc} compares the accuracy of {\proj} against a range of baselines. 
Notably, \proj achieves the highest average accuracy across all benchmarks, outperforming both single-agent and multi-agent methods. 
Specifically, it achieves state-of-the-art performance on \texttt{GSM8K}, 
 \texttt{MultiArith}, \texttt{SVAMP}, \texttt{AQuA}, and \texttt{HumanEval}, and ranks as the runner-up on \texttt{MMLU}.

The superior performance of \proj stems from multiple aspects. 
First, compared to single-agent methods, multi-agent collaboration allows agents to check and correct each other's reasoning, leading to higher accuracy. 
Second, compared to multi-agent methods with fixed routing structures, \proj generates task-specific multi-agent reasoning trajectories, enabling more adaptive and effective collaboration. 
Finally, compared to other learning-based routers, \proj is not constrained by manually defined anchor structures in {G-Designer} and fixed role distributions in {DyLAN} and {GPTSwarm}, offering greater flexibility and a larger routing solution space to explore optimal communication sequences.


\subsection{Robustness Evaluation}
\noindent\textbf{\proj is robust to malicious agents.}
An important advantage of learning-based multi-agent methods like \proj is their robustness to malicious agents. 
Through training, the router can learn to identify and downweight or bypass agents that provide misleading or harmful responses. 
To evaluate this, we conduct an experiment by injecting a malicious agent into the candidate agent pool. 
This agent is intentionally designed to produce consistently incorrect or distracting content.

\begin{figure*}[t]
  \centering
  \begin{minipage}[t]{0.61\textwidth}
    \centering
    \includegraphics[width=\linewidth]{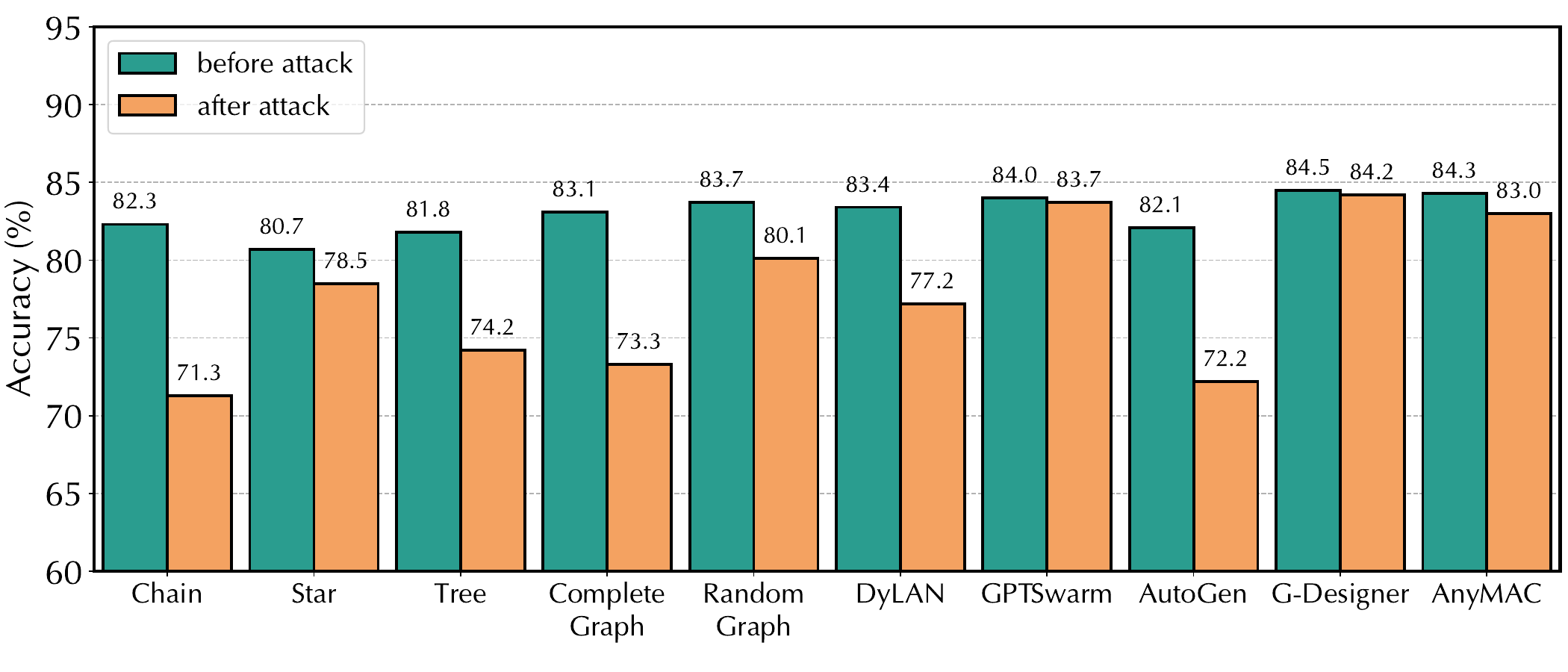}
    \caption{Robustness analysis of different methods on the \texttt{MMLU} dataset. We compare accuracy \textit{before} and \textit{after} the attack. Learning-based methods exhibit strong resilience to malicious agents.}
    \label{fig:robustness_mmlu}
  \end{minipage}
  \hfill
  \begin{minipage}[t]{0.37\textwidth}
    \centering
    \includegraphics[width=\linewidth]{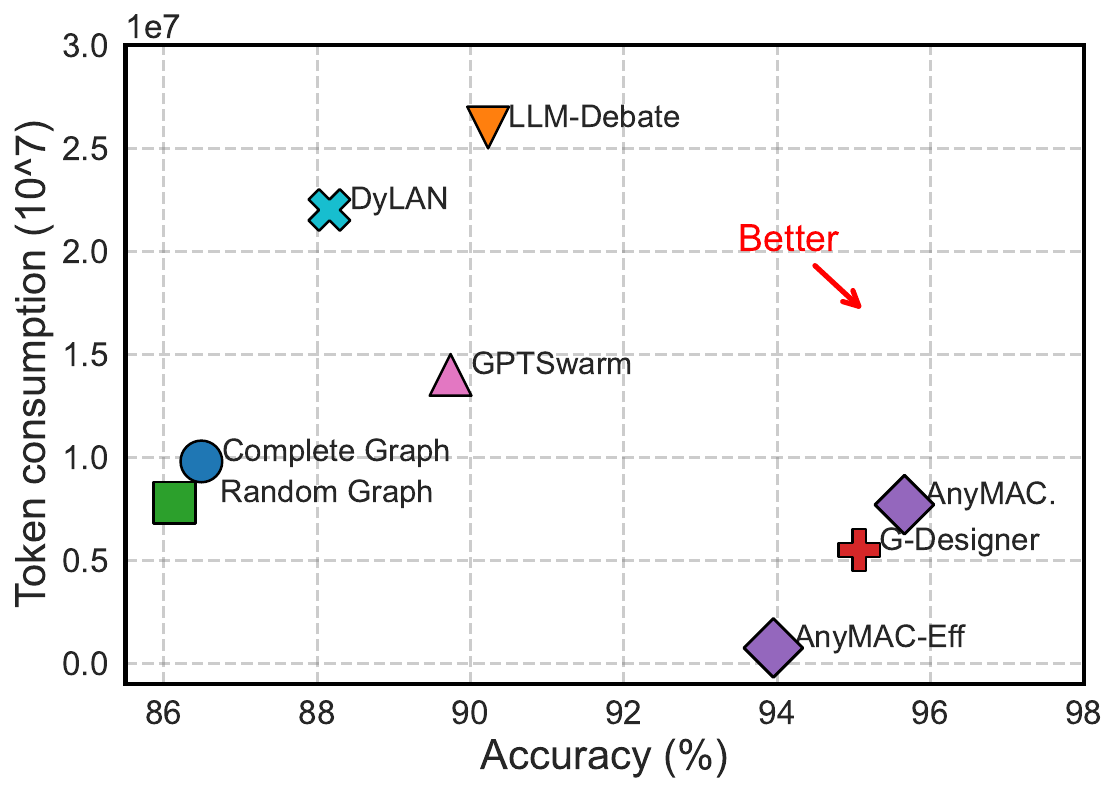}
    \caption{The performance and token consumption of various multi-agent communication topologies on \texttt{GSM8K}.}
    \label{fig:gsm8k_tradeoffs}
  \end{minipage}
\end{figure*}

Figure~\ref{fig:robustness_mmlu} shows the robustness evaluation results on \texttt{MMLU} dataset. 
We observe that other methods exhibit a significant accuracy drop (up to -11.0\%) when a malicious agent is present. 
In contrast, \proj maintains high accuracy (-1.3\%), highlighting its robustness under adversarial conditions.
We also observe that other learning-based baselines such as {GPTSwarm} and {G-Designer} exhibit similar robustness, indicating that learned routing policies generally confer a degree of resilience against adversarial inputs.


\subsection{Efficiency Evaluation}
\noindent\textbf{\proj is efficient.}
Beyond accuracy, token consumption is also a critical factor that affects practicality, as it reflects the actual cost incurred by users when solving tasks. 
Figure~\ref{fig:gsm8k_tradeoffs} compares the trade-off between prompt token usage and accuracy across different methods on the \texttt{GSM8K} dataset. 
Notably, {\proj} achieves the highest accuracy but with relatively higher token usage. 
In contrast, {\proj-Eff} sacrifices a small amount of accuracy in exchange for significantly improved efficiency, achieving the lowest token consumption among all methods. 
Compared to the most efficient baseline {G-Designer}, {\proj-Eff} reduces prompt token usage by 5$\times$, while maintaining competitive performance.
This demonstrates that our method is highly flexible: it can be configured to prioritize either accuracy or efficiency, and achieves state-of-the-art performance in both aspects.

\subsection{Qualitative Case Studies}

We further conduct qualitative analyses to evaluate the adaptability and behavior of \proj. 
First, we investigate whether \proj can adapt to different types of questions using the \texttt{MMLU} dataset. 
Figure~\ref{fig:qual_mmlu} in Appendix shows the routing results of {\proj} on two different questions. 
We observe that \proj selects different roles, showing its ability to adapt based on the question.

Furthermore, we also compare {\proj} and {\proj-Eff} using the same question from \texttt{GSM8K} to assess the trade-off between accuracy and efficiency. 
Figure~\ref{fig:qual_gsm8k} in Appendix shows that {\proj} selects more context and produces the correct answer, while {\proj-Eff} uses less context and fails. 
This shows that more computation can lead to improved accuracy.

\subsection{Ablation Studies}

In this subsection, we perform ablations on {\proj} and conduct experiments on the \texttt{GSM8K} dataset to evaluate the effectiveness of each module in our design. We consider the following variants: 
(1) Removing task adaptiveness. We replace the query embedding with a single learnable vector shared across all queries.
(2) Removing the Next-Agent Prediction (NAP) module. We replace the learned agent selection with a random choice at each step. 
(3) Removing the Next-Context Selection (NCS) module. We replace the context selection mechanism with random sampling from the context pool with 50\% probability for each response. 
The results are shown in Table~\ref{tab:ablation}. We observe that removing any of these components leads to an accuracy drop, highlighting their importance.

\begin{table}[!t]
    \centering
    \caption{Ablation study on {\proj} evaluated on the \texttt{GSM8K} dataset.}
    \label{tab:ablation}
    \resizebox{1\linewidth}{!}{
    \begin{tabular}{l|c}
        \toprule
        Variant & Accuracy (\%) \\
        \midrule
        {\proj} (Full Model) & 95.66 \\
        \midrule
        w/o Task Embed. (Task Adaptive) & 93.87 \\
        w/o Next-Agent Prediction (NAP) & 94.75 \\
        w/o Next-Context Selection (NCS) & 94.35 \\
        \bottomrule
    \end{tabular}
    }
\vspace{-1em}
\end{table}




\section{Conclusion}
\label{conclusion}

In this work, we investigate multi-agent collaboration through the lens of communication topology design. We propose a novel sequential formulation that dynamically constructs communication topologies by predicting the next agent at each step. Our framework introduces two key components: {Next-Agent Prediction} for flexible, task-aware role allocation, and {Next-Context Selection} for globally informed information routing. Together, these components enable a broader solution space than traditional graph structures, supporting adaptive, efficient, and robust multi-agent reasoning.
Extensive experiments across multiple benchmarks show that our approach consistently outperforms existing methods in both accuracy and efficiency. We believe this sequential communication protocol opens new directions for future research in scalable and generalizable multi-agent LLM systems.

\section{Limitations}
\label{sec:limit}

While \proj achieves the highest overall accuracy, its performance on the \texttt{HumanEval}, a code generation dataset, is slightly below the state-of-the-art as shown in Table~\ref{tab:main_acc}.
After investigating the reason, we find that \proj tends to select overly long contexts, which may overwhelm the LLM, leading to incorrect code generation. 

We believe the above issue is caused by an insufficient number of training samples (1,000) for reinforcement learning (RL), which may lead to convergence to a suboptimal solution. 
We hypothesize that this can be mitigated by scaling up RL training.
However, scaling RL sampling data for LLMs is both computationally and financially expensive in practice. 
Therefore, we leave this as future work and plan to explore it using smaller language models, which offer significantly lower data collection costs.
Moreover, smaller language models may benefit more from multi-agent collaboration due to their weaker individual capabilities.

\section{Ethical Consideration}
This work builds upon large language models (LLMs) for multi-agent collaboration and reasoning. All models and datasets used in our experiments are publicly available and widely adopted in the research community. We do not introduce any new data collection, nor do we engage in sensitive information.
During our experiments, we did not observe any explicit ethical concerns, harmful behaviors, or misuse cases. Nevertheless, we acknowledge that LLM-based systems, especially when deployed in multi-agent configurations, can potentially amplify biases or generate misleading information if not properly controlled. Our method focuses on improving multi-agent communication effectiveness and efficiency and does not alter the core generative behavior of the underlying LLMs.

\section*{Acknowledgements}
This work is supported in part by the National Science Foundation under grants (IIS-2006844, IIS-2144209,
IIS-2223769, CNS-2154962, BCS-2228534, and CMMI2411248), the Commonwealth Cyber Initiative Awards
under grants (VV-1Q24-011, VV-1Q25-004), and the research gift funding from Netflix and Snap.
\bibliography{ref}

\medskip

\clearpage
\appendix

\section{RL Training Details}
\label{sec:training}

\subsection{Exploration and Exploitation}

An important challenge in reinforcement learning is balancing the trade-off between \textit{Exploration} and \textit{Exploitation}. 
Exploration refers to discovering new possibilities by sampling uncertain actions, while exploitation focuses on selecting the best-known decision based on current knowledge.

In our framework, we encourage exploration by injecting noise into the output decision logits during training. 
By tuning the relative magnitude of the noise and decision score, we can effectively balance exploration and exploitation.

\paragraph{Next agent prediction (NAP).} For NAP, it is essential to balance the compatibility scores $s_i$ and the injected Gumbel noise (\Cref{sec:nap}). 
To achieve this, we first compute the mean and standard deviation of compatibility scores $s_i$ across $N$ roles: 
\begin{equation}
\mu = \frac{1}{N} \sum_{i=1}^{N} s_i', \quad 
\sigma = \sqrt{ \frac{1}{N} \sum_{i=1}^{N} (s_i' - \mu)^2 }.
\end{equation}
We then normalize the scores to have zero mean and scale them to have a standard deviation of \( \alpha \):
\begin{equation}\label{eq:nap}
s_i = \alpha \cdot \frac{s_i' - \mu}{\sigma}.
\end{equation}
By tuning \( \alpha \), we can control the determinism of the routing behavior: higher values of \( \alpha \) lead to more deterministic (exploitation-oriented) decisions, while lower values encourage exploration.

Empirically, we set \( \alpha = 1.5 \), as it provides a good balance between exploration and exploitation.

\paragraph{Next context selection (NCS).}
We apply a similar scaling strategy to the cosine similarities in NCS. 
Specifically, each similarity score \( c_j \) is multiplied by a scaling factor \( \beta \) to obtain \( c_j' \):
\begin{equation}
c_j' = c_j, \quad c_j' \in [-\beta, \beta],
\end{equation}
which is passed through a sigmoid function (\Cref{eqn:sigmoid}) to compute the sampling probability \( g_j \). 

Note that contexts are selected based on \( g_j \) using Bernoulli sampling. 
By adjusting \( \beta \), we control the determinism of context selection: a larger \( \beta \) makes the selection more deterministic (for exploitation), while a smaller \( \beta \) promotes exploration.
Empirically, \( \beta \) is set to 3, as it provides a good balance between exploration and exploitation.

\subsection{Reward Normalization}
\label{sec:rw_norm} 
As discussed in \Cref{sec:rl}, directly comparing rewards across different questions is inappropriate, as their difficulty levels may vary. 
A high reward on an easy question may not reflect the effectiveness of the routing strategy but rather the simplicity of the task itself. 
To address this, we adopt a standard reward normalization trick~\citep{gu2016continuous, schulman2017proximal}, which normalizes the reward of each trajectory based on a baseline estimated from the same question:
\begin{equation}
A_t = \frac{r_t - \mu}{\sigma}, \quad
\mu = \frac{1}{T} \sum_{k=1}^{T} r_k,
\end{equation}
\begin{equation}
\sigma = \sqrt{\frac{1}{T} \sum_{k=1}^{T} (r_k - \mu)^2}.
\end{equation}
Here, \( r_t \) denotes the reward of the current trajectory, and \( T \) is the number of sampled trajectories for the same question. 
Each \( r_k \) represents the reward of the \( k \)-th sampled trajectory.
The resulting \( A_t \) is the advantage value of the current trajectory in policy gradient optimization (\Cref{eqn:pg}), reflecting how much better (or worse) the current trajectory performs compared to the average baseline.

This normalization trick mitigates the impact of question difficulty and enables the reinforcement learning process to treat rewards from different queries more fairly and consistently.

\clearpage
\begin{figure*}[t]
    \centering
    \begin{minipage}{0.83\linewidth}
        \centering
        \includegraphics[width=\linewidth]{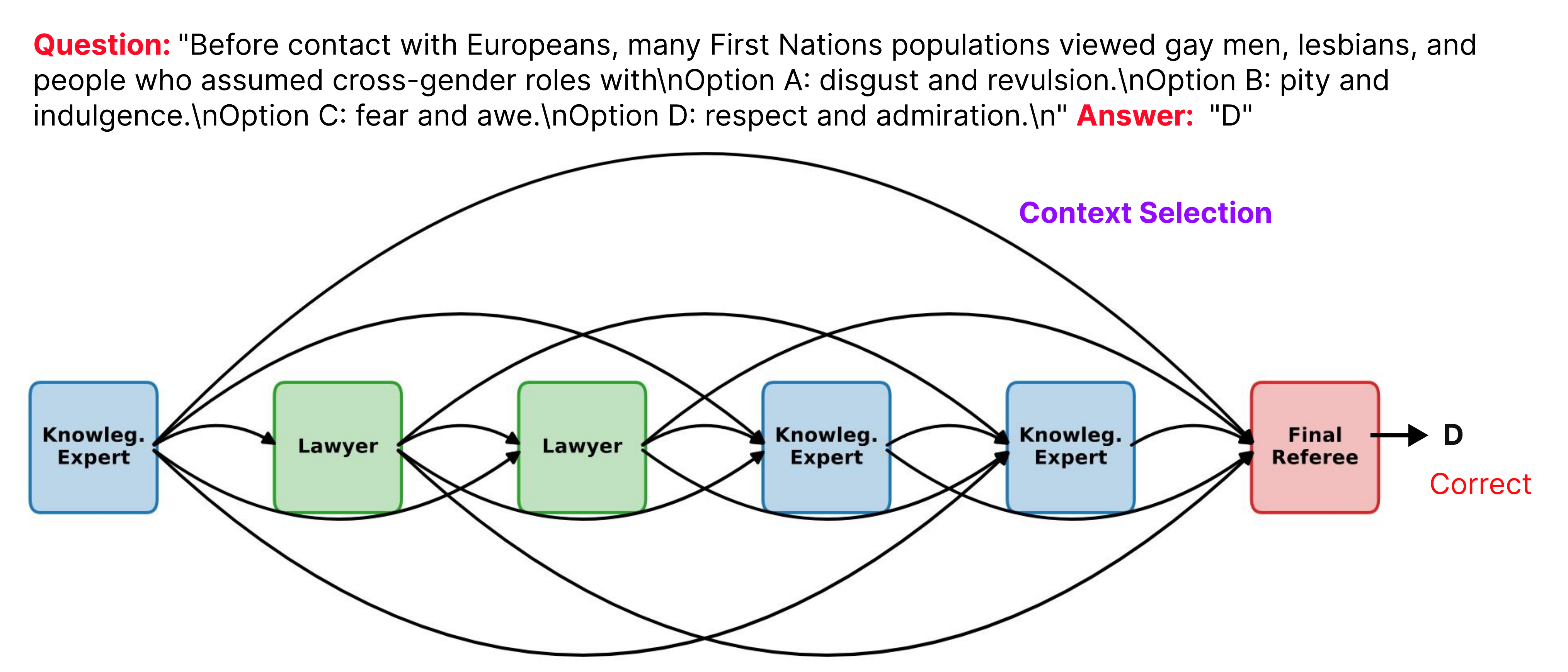}
        \caption*{(a) Routing results and final answer of \mode{\proj} for Question 1 in \texttt{MMLU}.}
        \label{fig:mmlu_qual1}
    \end{minipage}

    \vspace{1em}  

    \begin{minipage}{0.83\linewidth}
        \centering
        \includegraphics[width=\linewidth]{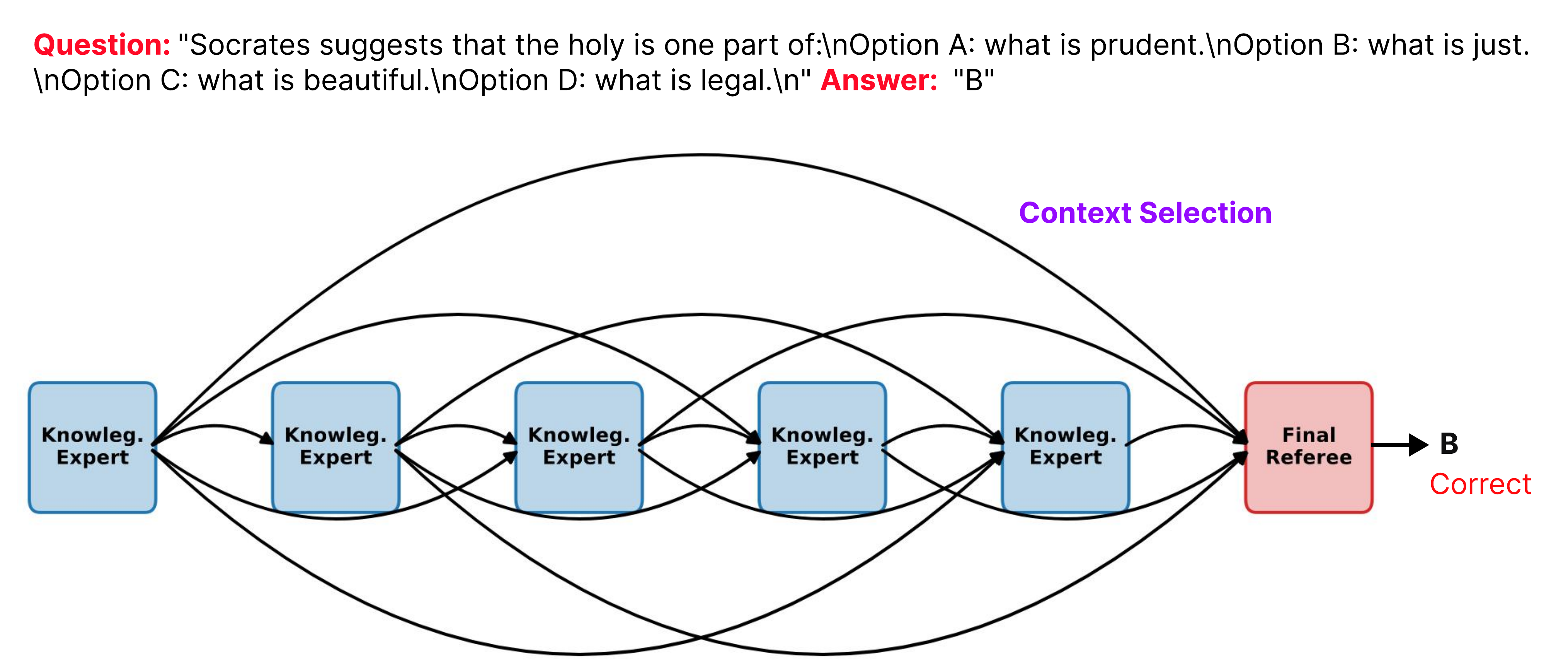}
        \caption*{(b) Routing results and final answer of \mode{\proj} for Question 2 in \texttt{MMLU}.}
        \label{fig:mmlu_qual2}
    \end{minipage}

    \caption{
    Qualitative illustration of {\proj}'s routing decisions on two different questions. 
    In (a), the model selects a combination of \textit{Lawyer} and \textit{Knowledgeable Expert} for Question 1. 
    In (b), it assigns all agents as \textit{Knowledgeable Experts} for Question 2.
    }

    \label{fig:qual_mmlu}
\end{figure*}

\begin{figure*}[t]
    \centering
    \includegraphics[width=0.83\linewidth]{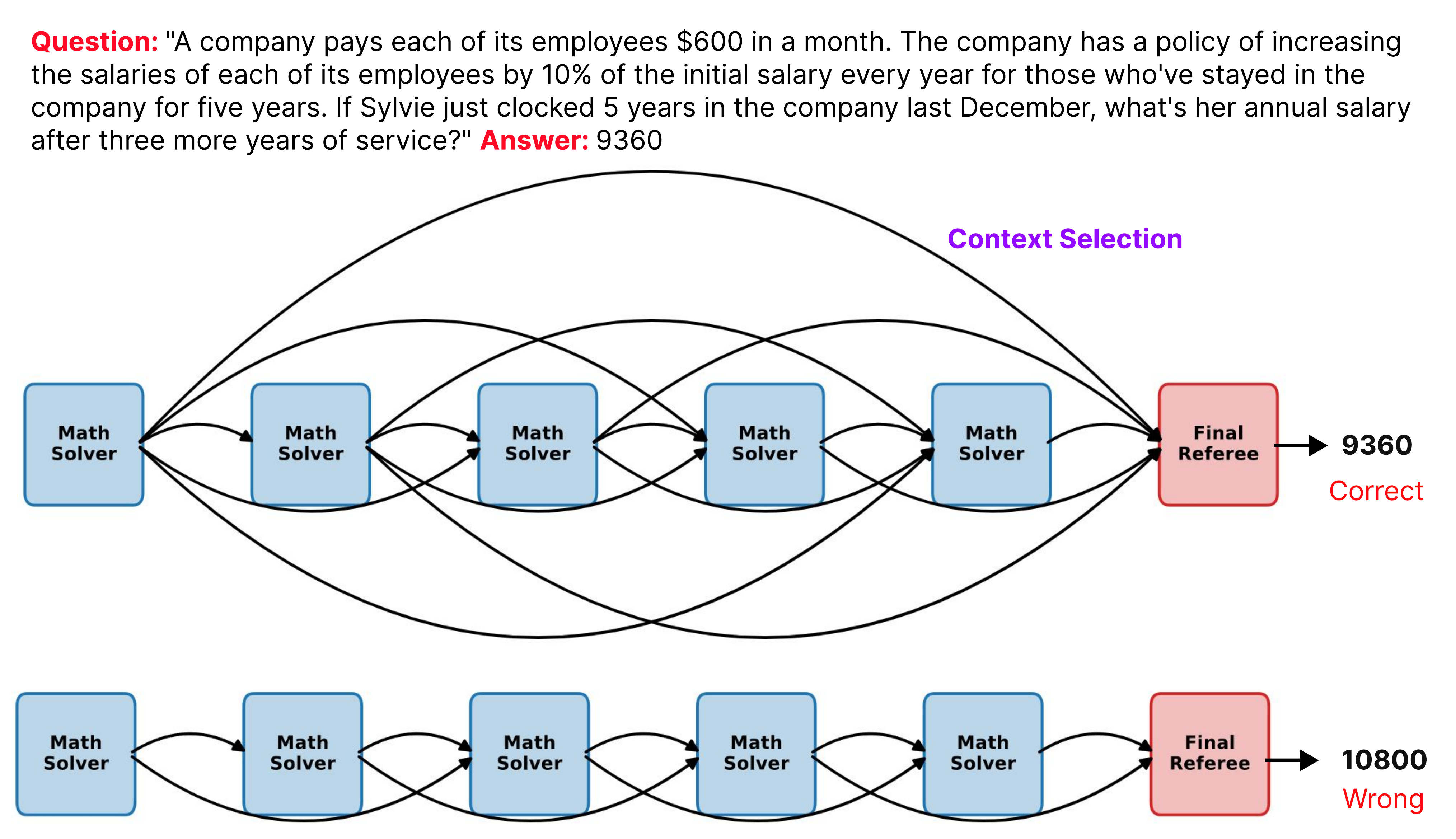}
\caption{
Qualitative comparison between \mode{\proj} (top) and \mode{\proj-Eff.} (bottom) on the same question. 
They use different computation budgets, and the increased computation in \mode{\proj} leads to the correct answer.
}
    \label{fig:qual_gsm8k}
\end{figure*}

\end{document}